\documentclass{article}

\PassOptionsToPackage{numbers, sort&compress}{natbib}
\usepackage[main,final]{neurips_2026}

\usepackage[utf8]{inputenc} 
\usepackage[T1]{fontenc}    
\usepackage{hyperref}       
\usepackage{url}            
\usepackage{booktabs}       
\usepackage{amsfonts}       
\usepackage{nicefrac}       
\usepackage{microtype}      
\usepackage{xcolor}         

\usepackage{amsmath, amssymb, amsfonts}
\usepackage{bm}
\usepackage{graphicx}
\usepackage{booktabs}
\usepackage{multirow}
\usepackage{hyperref}
\usepackage{cleveref}
\usepackage{wrapfig}
\usepackage{array}
\usepackage{color}

\newcommand{\dlt}[1]{\rlap{\textsubscript{$\uparrow$#1}}}
\newcolumntype{G}{c<{\phantom{\textsubscript{$\uparrow$0.0}}}}
\newcommand{\std}[1]{\rlap{\textsubscript{$\pm$#1}}}
\newcolumntype{Q}{c<{\phantom{\textsubscript{$\pm$0.00}}}}

\usepackage{subcaption}
\usepackage{makecell}
\usepackage{threeparttable}
\title{Fast-Slow Evolutionary Occupancy Prediction \\ via Controlled Dynamics}

\author{
\textbf{Runhe Yang}$^{1,*}$
\quad
\textbf{Zhiyuan Zhou}$^{1,*}$
\quad
\textbf{Yuxiang Yan}$^{1}$
\quad
\textbf{Xuetong Yang}$^{1}$
\quad
\textbf{Yikai Pan}$^{1}$
\\
\textbf{Jiacheng Tang}$^{1}$
\quad
\textbf{Zhuolin He}$^{1,2}$
\quad
\textbf{Jianhua Han}$^{2}$
\quad
\textbf{Hang Xu}$^{2}$
\quad
\textbf{Jian Pu}$^{1,\dagger}$
\\[3pt]
$^{1}$Fudan University
\qquad
$^{2}$Yinwang Intelligent Technology
\\
\texttt{\{rhyang26,zhouzy25\}@m.fudan.edu.cn}
\quad
\texttt{jianpu@fudan.edu.cn}
}

\begin{document}

\maketitle
\begingroup
\renewcommand{\thefootnote}{\fnsymbol{footnote}}
\footnotetext[1]{These authors contributed equally to this work.}
\footnotetext[2]{Corresponding author.}
\endgroup

\begin{abstract}
3D semantic occupancy prediction provides dense geometric and semantic scene representations for autonomous driving, where both high accuracy and low response latency are crucial for safe downstream forecasting and planning. Existing methods usually exhibit different accuracy-latency characteristics. Performance-oriented models provide stronger geometric and semantic predictions, while deployment-friendly models provide faster responses. This motivates EvoOcc, a fast-slow evolutionary occupancy prediction framework that jointly exploits their complementary strengths to improve the accuracy-latency balance without changing the upstream architectures. EvoOcc models evolutionary occupancy prediction as a continuous time controlled dynamical system, enabling flexible adaptation to variable slow system delays and irregular fast system arrivals. The framework further decomposes the evolution process into dual state evolution where ego state evolution compensates for the deterministic coordinate shifts and scene state evolution models the genuine geometric and semantic scene changes. Extensive experiments on three different fast-slow system pairs and irregular prediction arrivals show that EvoOcc consistently improves over fast system baselines while maintaining much lower latency than the slow system, indicating a favorable accuracy-latency balance and stable behavior under practical inference situations.

\end{abstract}

\section{Introduction}
\label{intro}

3D semantic occupancy prediction~\cite{li2023fbocc,tian2023occ3d} has emerged as an important scene representation for autonomous driving~\cite{tong2023scene}. It encodes the surrounding environment as a dense 3D volume with both geometric occupancy and semantic labels, providing cues for downstream motion forecasting~\cite{tang2024hpnet,zhang2024demo} and planning~\cite{zheng2024occworld,tang2026causalvad}. However, practical deployment requires not only high prediction accuracy but also low response latency. An occupancy prediction can quickly become outdated as the ego vehicle and surrounding agents move, potentially leading to unsafe or overly conservative planning decisions~\cite{yang2022real,li2020towards,tang2026decoupling}. These considerations make it essential to achieve a favorable balance between accuracy and latency for practical 3D semantic occupancy prediction.

\begin{figure*}[t]
  \centering
  \includegraphics[width=.95\textwidth]{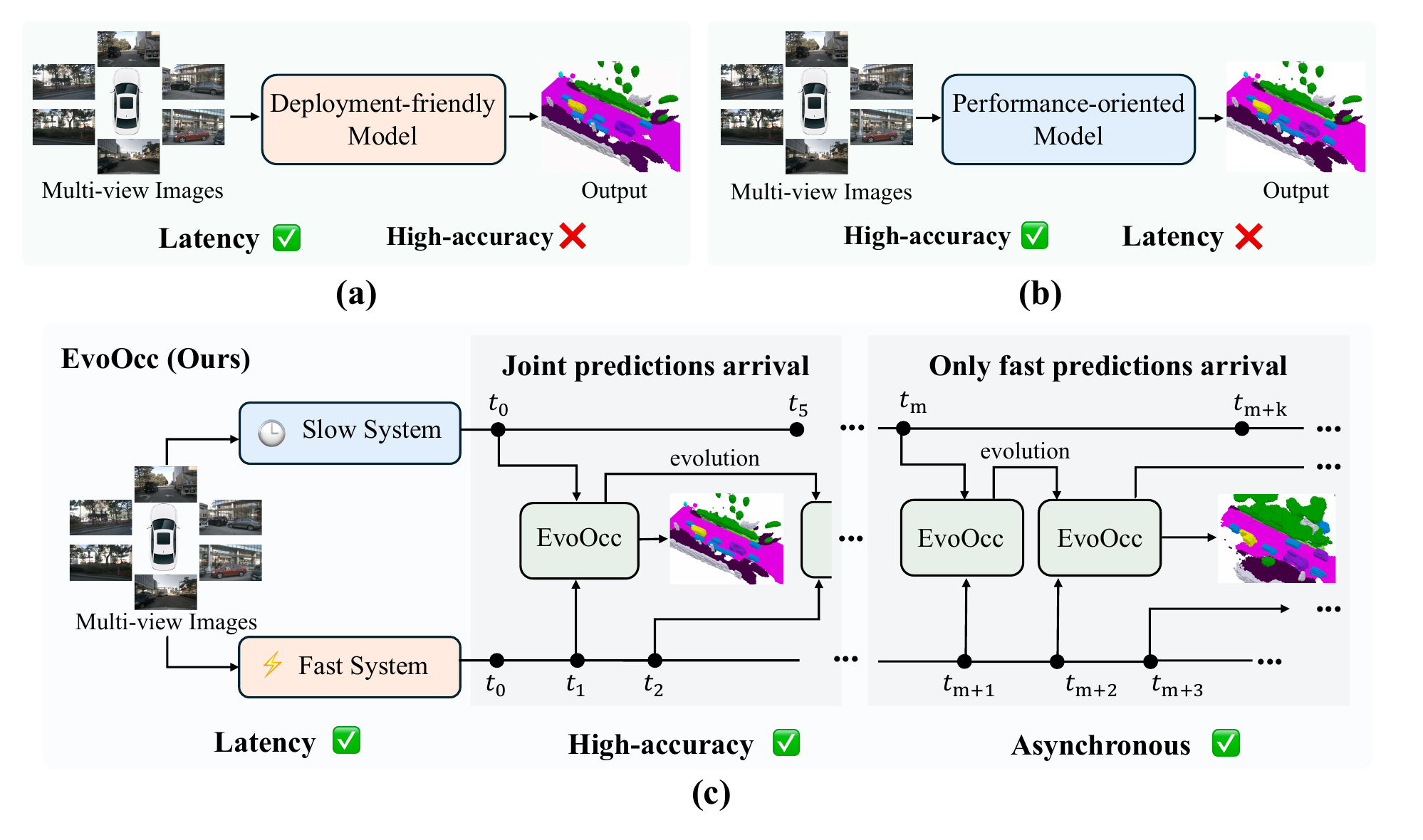}
  \caption{
\textbf{Overview of fast-slow evolutionary occupancy prediction.}
(a) Deployment-friendly models provide timely occupancy predictions but sacrifice geometric and semantic accuracy.
(b) Performance-oriented models are more accurate, yet their delayed outputs may become stale for real-time perception.
(c) EvoOcc exploits high-quality information from the slow system without sacrificing the responsiveness of the fast system by switching between two operational modes.
}
  \label{fig:Fig1}
\end{figure*}

Existing occupancy prediction systems have made various attempts to balance prediction accuracy and response latency. As illustrated in Fig.~\ref{fig:Fig1}, we categorize them into two types: \textit{deployment-friendly models} and \textit{performance-oriented models}~\cite{zhang2026vision}. Deployment-friendly models~\cite{wang2024opus,hou2024fastocc} adopt lightweight designs and produce low-latency predictions for the current scene, but their limited model capacity often leads to inferior geometric and semantic accuracy. In contrast, performance-oriented models~\cite{li2023fbocc,shi2025odg} use stronger architectures to obtain higher accuracy, yet their larger inference cost delays the output and makes it stale when the scene has already evolved. These properties make deployment-friendly models and performance-oriented models naturally complementary. Such complementarity suggests the potential to exploit high-quality information from performance-oriented models while preserving deployment-friendly responsiveness under asynchronous prediction arrival.

Motivated by the complementarity, we propose \textit{EvoOcc}, a fast-slow evolutionary occupancy prediction framework to improve both prediction accuracy and response latency. 
EvoOcc considers a realistic deployment scenario in which predictions of the fast and slow systems arrive asynchronously due to their different inference costs. 
To handle variable slow system delays and irregular fast system frame intervals, we formulate state evolution as a continuous time controlled dynamical system. This formulation allows EvoOcc to seamlessly adapt to asynchronous inputs by switching between two operational modes, as shown in Fig.~\ref{fig:Fig1}(c). 
Recognizing that variations in occupancy prediction arise from both ego vehicle motion and genuine scene changes~\cite{he2026dynamicvggt}, we decouple the evolution process into dual state evolution: ego state evolution and scene state evolution. 
Under joint predictions arrival, EvoOcc utilizes the high-quality slow system predictions to calibrate a robust historical anchor. When only fast predictions arrive, we design a dual state evolution to evolve the state carrying historical slow system information. Deterministic coordinate shifts induced by ego movement are compensated through ego state evolution, while genuine geometric and semantic scene changes are captured through scene state evolution. By continuously evolving historical anchors to the present, EvoOcc achieves a favorable accuracy-latency trade-off.

We conduct extensive experiments in both offline and realistic streaming scenarios across three fast-slow system pairs, including two homogeneous pairs and one heterogeneous pair. The results show that EvoOcc consistently outperforms the fast system while maintaining much lower latency than the slow system, demonstrating that delayed high-quality predictions can effectively benefit real-time occupancy prediction when properly evolved. 
We further evaluate EvoOcc under sparse slow system invocations and irregular fast system arrivals.
Our findings confirm the stability of the proposed framework and validate its favorable balance between accuracy and latency.

Our contributions are summarized as follows: (i) We introduce \textit{EvoOcc}, a fast-slow evolutionary framework for occupancy prediction, which jointly exploits the complementary strengths of deployment-friendly and performance-oriented occupancy predictors to achieve a more favorable accuracy-latency balance. (ii) We formulate the framework as a continuous time controlled dynamical system and design a dual state evolution to model ego vehicle motion and genuine geometric and semantic scene changes. (iii) Extensive experiments on the Occ3D-nuScenes benchmark under three different fast-slow system pairs show that EvoOcc improves fast system predictions while maintaining much lower latency than slow system baselines and remains stable under irregular arrival conditions.

\section{Related Work}
\label{related_works}

\subsection{3D Semantic Occupancy Prediction}

3D semantic occupancy prediction represents the driving scene as a voxelized semantic volume with occupancy and semantic labels~\cite{wang2023openoccupancy,wang2025uniocc,yan2024pointssc}. MonoScene~\cite{cao2022monoscene} lifts monocular features into 3D for semantic scene completion, while recent surround-view methods~\cite{huang2023tri,li2023voxformer} improve dense occupancy modeling through multi-camera volumetric features in SurroundOcc~\cite{wei2023surroundocc}, forward-backward view transformation in FB-Occ~\cite{li2023fbocc}, and adaptive lifting in ALOcc~\cite{chen2025alocc}. However, dense voxel modeling is computationally expensive. Efficient alternatives~\cite{huang2024gaussianformer,huang2025gaussianformer} thus explore compact or sparse representations~\cite{zhang2023occformer}: FastOcc~\cite{hou2024fastocc} uses a lightweight BEV design, SparseOcc~\cite{liu2024fully} adopts sparse 3D computation, and OPUS~\cite{wang2024opus} predicts occupancy via sparse set queries. 
Different from prior work that improves a single prediction framework, we view these methods as complementary fast and slow systems and propose fast-slow evolutionary occupancy prediction, which evolves residual information between slow and fast logits so that high-quality slow predictions can complement timely fast predictions, leading to a more favorable accuracy-latency balance.

\subsection{Fast-Slow Architectures}

Fast-slow architectures are motivated by Kahneman's view~\cite{kahneman2011thinking}, where fast processing is immediate and lightweight while slow processing is more deliberate and capable. In multimodal agents, fast pathways are commonly used for direct visual perception, while slow pathways are invoked for more complex reasoning or decision making~\cite{sun2025visual}. Recent vision-language-action and robotic systems further combine high-frequency action generation with low-frequency reasoning or hierarchical planning, where the slow system provides cognitive guidance and the fast system enables responsive control~\cite{chen2025fast,zhang2024hirt}. These works mainly study fast-slow coordination for reasoning, planning, or control~\cite{li2025system,xiao2025fast}. In contrast, EvoOcc focuses on 3D occupancy prediction in autonomous driving. EvoOcc treats existing fast and slow occupancy predictors as frozen asynchronous logit sources and evolves their complementary information without modifying the upstream architectures.

\subsection{Neural Differential Equations}

Neural differential equations model hidden representations through continuous time dynamics. Neural ODEs~\cite{chen2018neural} parameterize state derivatives with neural networks and have been extended to irregularly sampled sequences by latent ODE~\cite{rubanova2019latent} and GRU-ODE-Bayes models~\cite{de2019gru}. StreamingFlow~\cite{shi2024streamingflow} applies neural ODEs to asynchronous multi-modal streams for BEV occupancy forecasting at arbitrary timestamps. In standard Neural ODE formulations, once the initial state is given, the continuous trajectory is governed by the learned dynamics rather than continuously driven by incoming observations~\cite{morrill2021neural}. Neural controlled differential equations~\cite{kidger2020neural} instead treat observations as continuous control paths, enabling states to evolve under irregular external inputs. GraphNCDE~\cite{choi2022graph} further combines NCDEs with graph neural networks for spatio-temporal traffic forecasting. EvoOcc follows this controlled-dynamics perspective for fast-slow evolutionary occupancy prediction, using the fast system prediction stream as a control signal to evolve delayed residual information.

\section{Method}
\label{sec:method}

\begin{figure*}[t]
  \centering
  \includegraphics[width=.98\textwidth]{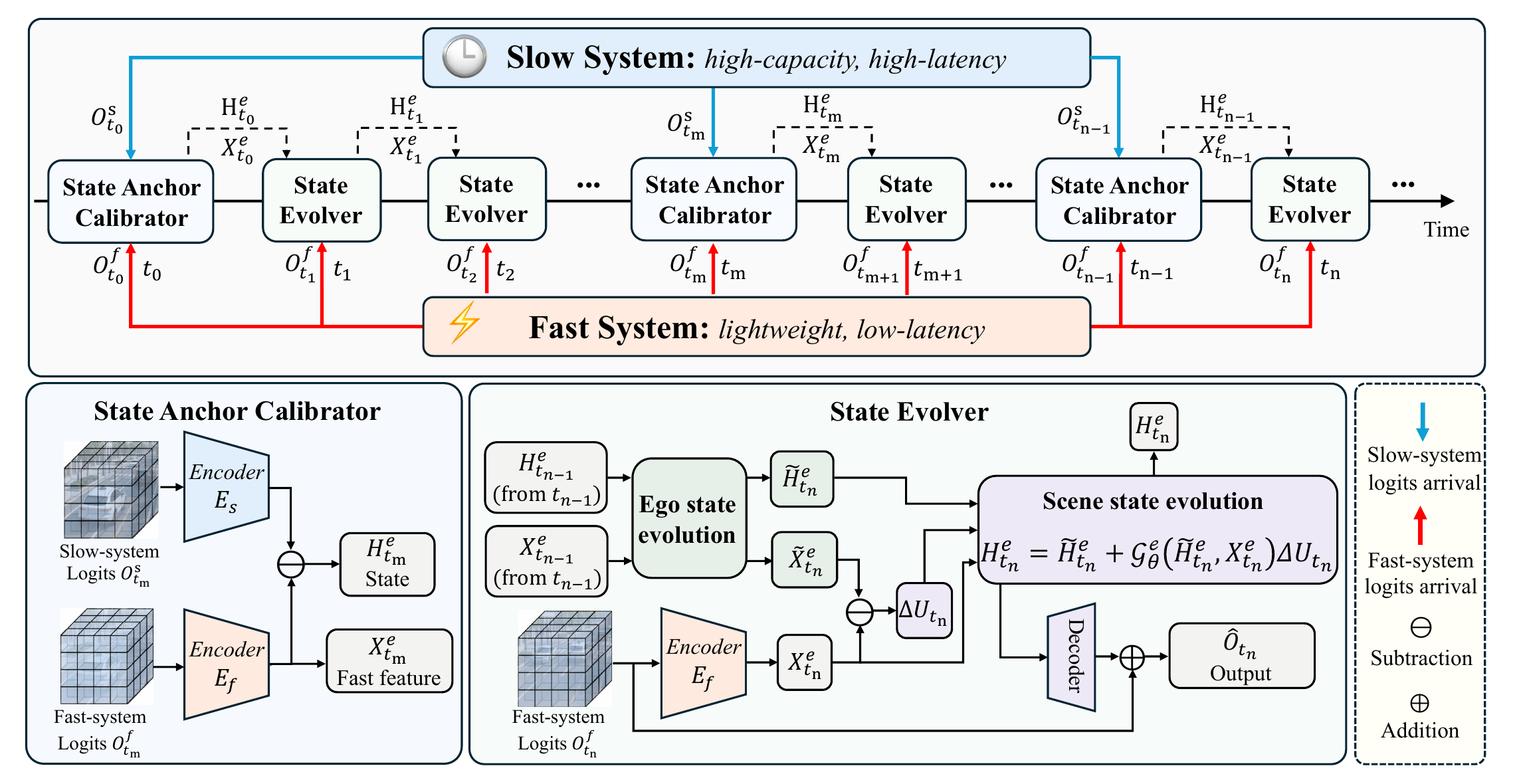}
  \caption{
  \textbf{Main framework of EvoOcc.} Along the timeline, EvoOcc switches between two operational modes depending on which logits arrive at the current timestamp. When slow system logits arrive, EvoOcc uses them to calibrate the state anchor. When only the fast system logits arrive, EvoOcc evolves the previous state by ego state evolution and scene state evolution, then decodes it as a residual correction to the current fast logits.
  }
  \label{fig:fig2}
\end{figure*}

To jointly exploit the strength of fast and slow systems without changing their architectures, we propose \textit{EvoOcc}, a fast-slow evolutionary occupancy prediction framework that balances accuracy and latency for real-time perception. As illustrated in Fig.~\ref{fig:fig2}, we model the dynamic process of evolutionary occupancy prediction as a continuous time controlled dynamical system to handle irregular fast system intervals and variable slow system delays. This formulation motivates a dual state evolution design to model the occupancy semantic changes caused by ego movement and scene changes. Finally, we train EvoOcc with a multi-task objective that combines voxel-wise semantic supervision with ray-guided geometric consistency to improve surface localization.

\subsection{Problem Formulation}
\label{sec:method:problem}

We study fast-slow evolutionary occupancy prediction under asynchronous logits arrival, where a high-capacity slow system and a lightweight fast system produce their occupancy logits, denoted by $O^s$ and $O^f$, with different latencies and update frequencies. At the current timestamp $t_n$, the fast system provides timely occupancy logits $O^f_{t_n}$, while the latest available slow system output $O^s_{t_m}$ may correspond to an earlier timestamp $t_m \leq t_n$ due to its larger inference latency. 

Our task is to learn an evolutionary mapping $\mathcal{F}_{\theta}: \{O_{t_m}^s, \mathcal{O}_{t_m:t_{n}}^f\} \to \hat{O}_{t_{n}}$, where 
$\mathcal{O}_{t_m:t_n}^f$ denotes the fast logits observed between $t_m$ and $t_n$ and $\hat{O}_{t_{n}}$ denotes the refined occupancy prediction. Specifically, evolutionary mapping uses the delayed slow logits and the intervening fast logits stream to compensate the gap between $O^s_{t_m}$ and $O^f_{t_n}$. By operating directly on logits, $\mathcal{F}_\theta$ remains agnostic to specific upstream architectures while ensuring real-time response with higher accuracy.

\subsection{Continuous Time State Dynamics}
\label{sec:method:dynamics}

In practical deployments, slow logits may arrive with variable delays, and fast observations may have irregular temporal intervals. This asynchrony makes fixed step discrete updates less suitable, since the elapsed time between observations is not constant. We therefore formulate state evolution as a continuous time controlled dynamical system, which allows the state to evolve over arbitrary temporal intervals driven by the fast logits stream. Specifically, the state to be evolved is a residual between the slow and fast system logits, encoding the complementary information including high-quality geometric and semantic details.


\paragraph{Dynamics at Fixed Physical Locations.} A direct temporal update in the ego frame entangles apparent coordinate shifts caused by ego motion with genuine scene changes. To define state evolution at fixed physical locations, we first use the world coordinate system as a reference for the dynamics. The state field $H_t^w$ evolves under the control of the encoded fast feature field $X_t^w$:
\begin{equation}
dH_t^w = \mathcal{G}_\theta^w(H_t^w, X_t^w) dX_t^w,
\label{eq:world_cde}
\end{equation}
where $\mathcal{G}_\theta^w$ is a convolutional local spatial operator, and the differential $d$ is taken with respect to time $t$. Eq.~\eqref{eq:world_cde} defines how the state changes at fixed physical locations according to temporal changes in the fast feature field. The convolutional operator further allows each local update to use neighboring voxel context, so the evolution is not restricted to isolated voxel wise changes.

\paragraph{Dynamics in Moving Ego Coordinates.} Fast and slow logits are predicted in ego coordinates. To use these inputs without modifying their original predictors, we express the same dynamics in the ego coordinate system. The corresponding ego coordinate fields are written as $H_t^e$ and $X_t^e$, and ego motion induces an apparent velocity field $u_t$ over the grid. The coordinate transformation gives
\begin{equation}
dH_t^w = dH_t^e + (u_t \cdot \nabla H_t^e)\,dt,\quad
dX_t^w = dX_t^e + (u_t \cdot \nabla X_t^e)\,dt,
\label{eq:ego_transport_identities}
\end{equation}
where the derivation is provided in Appendix~\ref{app:ego_derivation}. Eq.~\eqref{eq:ego_transport_identities} makes the mixed nature of ego frame updates explicit and shows that the variation measured at a fixed physical location can be expressed as the sum of two ego frame components: the temporal variation observed on the ego grid and the apparent transport caused by ego motion. Substituting Eq.~\eqref{eq:ego_transport_identities} into Eq.~\eqref{eq:world_cde} rewrites the world reference dynamics using ego coordinate quantities:
\begin{equation}
dH_t^e + (u_t \cdot \nabla H_t^e)\,dt
=
\mathcal G_\theta^e(H_t^e, X_t^e)
\Bigl(
dX_t^e + (u_t \cdot \nabla X_t^e)\,dt
\Bigr),
\label{eq:ego_transport_cde_expanded}
\end{equation}
where $\mathcal G_\theta^e$ denotes the local spatial operator used in ego coordinates. Eq.~\eqref{eq:ego_transport_cde_expanded} shows that the ego coordinate dynamics contain two coupled components: the left-hand side accounts for ego motion induced state transport, while the right-hand side interprets the driving fast feature change under the same coordinate transformation. For compact notation, we collect the motion compensated fast feature change into a control increment:
$dU_t := dX_t^e + (u_t \cdot \nabla X_t^e)\,dt $ .
This definition turns the fast feature variation into a control signal measured consistently with the moving ego frame. With this control increment, the ego coordinate dynamics become
\begin{equation}
\underbrace{dH_t^e}_{\text{ego variation}}+ \, \, \,
\underbrace{(u_t \cdot \nabla H_t^e)\,dt}_{\text{transport term}}
=
\underbrace{\mathcal G_\theta^e(H_t^e, X_t^e)\,dU_t}_{\text{controlled term}}.
\label{eq:ego_transport_cde}
\end{equation}
The transport term accounts for deterministic coordinate shifts induced by ego motion, while the controlled term models the remaining scene evolution driven by the fast logit stream. This structured design avoids treating ego motion and scene changes as a single entangled update. Ego state evolution first handles the deterministic coordinate shift using ego motion, and scene state evolution then evolves the state using the control signal from the fast logit stream.

\subsection{Dual State Evolution}
\label{sec:method:evolution}

To implement the continuous time state dynamics, EvoOcc contains a dual state evolution process. Inspired by operator-splitting ~\cite{strang1968construction,mclachlan2002splitting}, EvoOcc addresses these dynamics sequentially. EvoOcc first performs ego state evolution to compensate for coordinate shifts, followed by scene state evolution to capture environmental changes. When fresh slow system logits arrive, EvoOcc calibrates the state anchor to refresh the state with high-quality residual information.

\paragraph{State anchor calibrator.}
 
When fresh slow system logits $O^s_{t_m}$ become available, EvoOcc calibrates a new state anchor for future state evolution. Specifically, we encode the current slow and fast system logits and calibrate the state anchor as follows:
\begin{equation}
H^e_{t_m} = E_s(O^{s}_{t_m}) - E_f(O^{f}_{t_m}),
\label{eq:slow_refresh_hidden_state}
\end{equation}
where \(E_s\) denotes the slow-branch encoder and \(E_f\) denotes the fast-branch encoder. This calibrated anchor explicitly captures the geometric and semantic details provided by the slow system but absent from the fast system, thereby establishing a high-quality state anchor for subsequent state evolution.

\paragraph{Ego state evolution.}
When only fast system logits arrive at timestamp $t_n$, no fresh slow system logits are available to recalibrate the state anchor. We therefore evolve the historical anchor from the previous timestamp $t_{n-1}$. We first consider the deterministic coordinate shifts caused by ego movement. The ego motion from $t_{n-1}$ to $t_n$ induces a spatial warping  over the feature grid.
 We apply the operator $W_{\mathrm{ego}}$  to evolve both previous state anchor and fast features into the current ego grid, 
\begin{equation}
\tilde{H}_{t_n}^{e} = W_{\mathrm{ego}}(H^e_{t_{n-1}}),
\qquad
\tilde{X}_{t_n}^{e} = W_{\mathrm{ego}}(X^e_{t_{n-1}}),
\label{eq:ego-warp}
\end{equation}
where \(\tilde{H}_{t_n}^{e}\) denotes the evolved state anchor and \(\tilde{X}_{t_n}^{e}\) denotes the evolved fast system feature. By compensating for these coordinate shifts, this step removes the influence of ego motion, ensuring that the remaining occupancy variations are driven by genuine scene changes.

\paragraph{Scene state evolution.}
After ego state evolution, EvoOcc uses the discrepancy between the evolved fast system feature \(\tilde{X}_{t_n}^{e}\)  and the current fast system feature $X^e_{t_n}$ to characterize scene changes. Since both features are expressed in the same coordinate system, their difference $\Delta U_{t_n} = X^e_{t_n} - \tilde{X}_{t_n}^{e}$ measures the scene variation that cannot be explained by ego motion. Moreover, $\Delta U_{t_n}$ serves as a discrete approximation of the control increment \(dU_t\) in Eq.~\eqref{eq:ego_transport_cde} over \([t_{n-1}, t_n]\), as derived in Appendix~\ref{app:ego_derivation}. Given this control increment, EvoOcc discretizes the scene state evolution with a forward Euler step:
\begin{equation}
H^e_{t_n}
=
\tilde{H}_{t_n}^{e}
+
\mathcal G_\theta^e(\tilde{H}_{t_n}^{e}, X^e_{t_n})\,\Delta U_{t_n},
\label{eq:motion_decoupled_euler}
\end{equation}
where $H^e_{t_n}$ denotes the final state after dual state evolution. Finally, EvoOcc decodes $H^e_{t_n}$ into a residual correction for the current fast system logits, yielding the refined occupancy prediction.

\subsection{Training Objective}
\label{sec:method:instantiation_training}

EvoOcc is supervised at every timestamp for which occupancy ground truth is available. The loss function consists of a voxel-wise semantic loss and a geometric regularization term that constrains surface placement. As discussed in SparseOcc~\cite{liu2024fully}, voxel-level supervision alone may not sufficiently constrain surface placement, and can favor over-thickened or depth-shifted occupancy. To better constrain the first visible surface along each ray, we complement the semantic supervision with a ray-termination likelihood loss, motivated by ray-based consistency~\cite{tulsiani2017multi} and depth-supervised volume-rendering objectives~\cite{deng2022depth}. Along each ray $r$, the $i$-th sampled point is assigned a predicted free-space probability $p_{r,i}^{\mathrm{free}}$. We denote the occupied probability \(p_{r,i}\), the first-hit probability \(q_{r,i}\), and the ray-termination likelihood loss as
\begin{equation}
p_{r,i}=1-p_{r,i}^{\mathrm{free}},\quad
q_{r,i}=p_{r,i}{\textstyle\prod_{j<i}}(1-p_{r,j}),\quad
\mathcal L_{\mathrm{term}}
=-\log\!\left({\textstyle\sum_{i\in\mathcal W_r}}q_{r,i}\right),
\label{eq:ray_termination_loss}
\end{equation}
where \(\mathcal{W}_r\) is a small local window around the ground-truth surface hit location. The overall training loss \(\mathcal L\) is then the weighted sum of the semantic and geometric terms over all supervised timestamps:
\begin{equation}
\mathcal L = \sum_{k\in\mathcal T_{\mathrm{sup}}} \left( \mathcal L_{\mathrm{sem}}^{(k)} + \lambda_{\mathrm{term}}\mathcal L_{\mathrm{term}}^{(k)} \right),
\label{eq:total_training_objective}
\end{equation}
where \(\mathcal T_{\mathrm{sup}}\) is the supervised-timestamp set, and \(\mathcal L_{\mathrm{sem}}\) combines weighted focal~\cite{lin2017focal} and Lov\'asz~\cite{berman2018lovasz} losses to handle class imbalance and optimize an IoU surrogate.

\section{Experiments}
\label{experiments}

\subsection{Experimental Setup}
\label{experiments:4.1}

\paragraph{Dataset and metrics.}
We evaluate our EvoOcc on the validation split of Occ3D-nuScenes~\cite{tian2023occ3d}, which provides 1,000 annotated scenes for the nuScenes~\cite{caesar2020nuscenes} with the standard 700/150/150 train/val/test split. We report mIoU, mIoU\textsubscript{D}~\cite{chen2025alocc}, and RayIoU~\cite{liu2024fully}, where mIoU evaluates the 17 non-free semantic classes, mIoU\textsubscript{D} evaluates the subset of 8 dynamic foreground classes and RayIoU is averaged over distance thresholds of 1 m, 2 m, and 4 m.

\paragraph{Implementation details.}
We consider three different frozen fast-slow system pairs, including two homogeneous pairs and one heterogeneous pair: ALOcc-2D-mini/ ALOcc-3D~\cite{chen2025alocc}, OPUS-T~\cite{wang2024opus}/ OPUS-V2-L~\cite{wang2026opus}, and ALOcc-2D-mini/ OPUS-V2-L. We only train EvoOcc without modifying the original fast and slow system. 
During training, each sample takes the slow system prediction from 2.0\,s earlier and evolves the state to the current frame.
The model is trained for 20 epochs with batch size 2 on two NVIDIA RTX 5090 GPUs. We optimize the training process with AdamW~\cite{loshchilov2019decoupled} using a learning rate of $1.0\times10^{-4}$, weight decay 0.01, gradient clipping at max norm 5.0, and one warmup epoch. Unless otherwise specified, model training uses random seed 0. Besides, we maintain an exponential moving average with decay 0.999.

\begin{table}[t]
\centering
\caption{Results for EvoOcc in offline evolution.
Fast system predictions arrive every 0.5\,s, and each current timestamp is paired with its own slow system prediction from 2.0\,s earlier. The result of slow system evaluated at the current timestamp is reported only as a reference. Warped delayed slow applies only ego-motion warping to align the delayed slow predictions to the current timestamp. 
Values after $\uparrow$ denote absolute gains over the fast system.}
\label{tab:tab1}
\footnotesize
\setlength{\tabcolsep}{5pt}
\begin{tabular}{@{}ll*{6}{c}@{\hspace{10pt}}}
\toprule
Fast / Slow pair & Method & mIoU & mIoU\textsubscript{D}
 & RayIoU\textsubscript{1m} & RayIoU\textsubscript{2m} & RayIoU\textsubscript{4m} & RayIoU \\
\midrule
 & Fast System         & 33.4 & 31.3 & 32.9 & 40.1 & 44.8 & 39.3 \\
ALOcc-2D-mini~\cite{chen2025alocc} /
 & Slow System         & 38.3 & 35.6 & 37.7 & 44.7 & 48.8 & 43.7 \\
ALOcc-3D
 & Warped delayed slow & 26.5 & 20.0 & 25.1 & 30.5 & 34.4 & 30.0 \\
 & EvoOcc (ours)
 & 39.3\dlt{5.9} & 33.7\dlt{2.4} & 34.7\dlt{1.8}
 & 41.3\dlt{1.2} & 45.6\dlt{0.8} & 40.5\dlt{1.2} \\
\midrule
 & Fast System         & 33.2 & 28.7 & 31.7 & 39.2 & 44.3 & 38.4 \\
OPUS-T~\cite{wang2024opus} /
 & Slow System         & 38.6 & 34.0 & 37.9 & 44.9 & 49.1 & 44.0 \\
OPUS-V2-L~\cite{wang2026opus}
 & Warped delayed slow & 25.3 & 17.7 & 24.5 & 30.0 & 34.1 & 29.5 \\
 & EvoOcc (ours)
 & 36.9\dlt{3.7} & 31.6\dlt{2.8} & 33.3\dlt{1.6}
 & 40.3\dlt{1.1} & 44.8\dlt{0.5} & 39.5\dlt{1.1} \\
\midrule
 & Fast System         & 33.4 & 31.3 & 32.9 & 40.1 & 44.8 & 39.3 \\
ALOcc-2D-mini /
 & Slow System         & 38.6 & 34.0 & 37.9 & 44.9 & 49.1 & 44.0 \\
OPUS-V2-L
 & Warped delayed slow & 25.4 & 18.3 & 24.5 & 29.9 & 33.9 & 29.4 \\
 & EvoOcc (ours)
 & 39.1\dlt{5.7} & 33.6\dlt{2.3} & 34.8\dlt{1.9}
 & 41.9\dlt{1.8} & 46.6\dlt{1.8} & 41.1\dlt{1.8} \\
\bottomrule
\end{tabular}
\end{table}

\begin{figure*}[t]
  \centering
  \includegraphics[width=.98\textwidth]{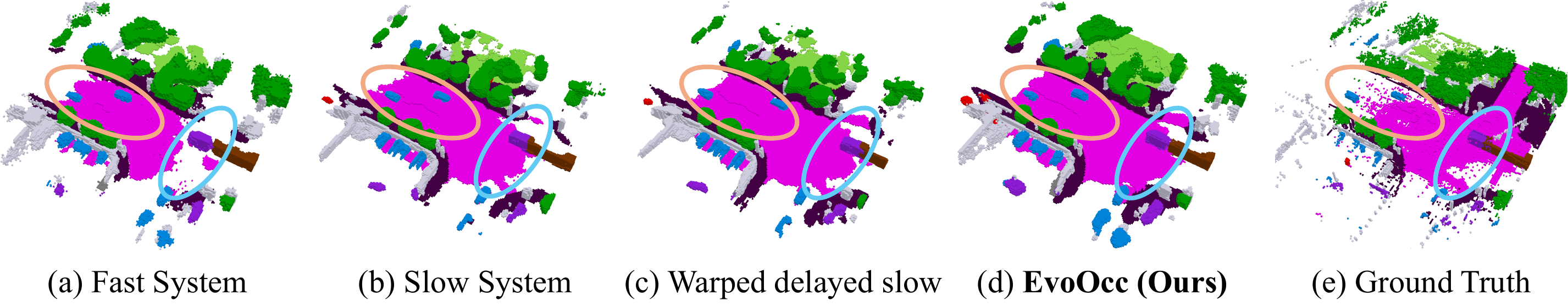}
  \caption{\textbf{Visualizations of 3D semantic occupancy predictions} on the OPUS-T / OPUS-V2-L pair. Orange and blue circles highlight representative dynamic-object and static-object regions.
  }
  \label{fig:vis_occ}
\end{figure*}

\subsection{Experimental Results}
\label{experiments:4.2}

\subsubsection{Experiments in Offline Evolution Scenarios}
We evaluate EvoOcc in an offline evolution scenario where fast system predictions arrive every 0.5\,s and each current timestamp is paired with a slow system prediction from 2.0\,s earlier. For boundary samples with shorter histories, we use the longest available history. As shown in Tab.~\ref{tab:tab1}, EvoOcc consistently improves over the timely fast system across all three fast-slow system pairs, with gains of 3.7-5.9 mIoU and 1.1-1.8 averaged RayIoU. Although this current time slow prediction is only an oracle reference and is unavailable in offline evolution, EvoOcc even surpasses the current-time slow system in mIoU for two of the three pairs. These results show that both voxel-level semantic quality and ray-based surface localization can benefit from our EvoOcc framework.

To further evaluate evolution of dynamic objects, we design a warped delayed slow baseline that applies only ego-motion warping~\cite{zhang2022beverse} to the delayed slow system prediction. As shown in Tab.~\ref{tab:tab1}, warped delayed slow performs substantially worse than the fast system on mIoU\textsubscript{D} due to its ignorance of genuine scene changes. In contrast, EvoOcc improves mIoU\textsubscript{D} by 2.3-2.8 points over the fast system and by 13.7-15.3 points over warped delayed slow. 
These results demonstrate that our dual state evolution uses the fast system prediction stream to evolve the historical anchor state, jointly handling coordinate shifts and genuine scene changes. The qualitative results in Fig.~\ref{fig:vis_occ} also illustrate that our EvoOcc benefits both static and dynamic objects. 
The orange and blue circles highlight cases where the fast system produces incomplete ground structures and warped delayed slow keeps stale object locations, while EvoOcc evolves these regions toward the current scene.

\begin{table}[t]
\centering
\caption{Results for EvoOcc in streaming evolution.
Fast system predictions arrive every 0.5\,s. A slow prediction sensed at $t_m$ becomes available at $t_n = t_m + 0.5$\,s, recalibrates the state anchor, and is evolved to the current timestamp $t_n$. Values after $\uparrow$ denote absolute gains over the fast system.}
\label{tab:tab2}
\footnotesize
\setlength{\tabcolsep}{5pt}
\begin{tabular}{@{}ll*{3}{G}@{}}
\toprule
Fast / Slow pair & Method & mIoU & mIoU\textsubscript{D} & RayIoU \\
\midrule
\multirow{3}{*}{\begin{tabular}{@{}l@{}}ALOcc-2D-mini /\\ ALOcc-3D\end{tabular}}
 & Fast System   & 33.4 & 31.3 & 39.3 \\
 & Slow System   & 38.3 & 35.6 & 43.7 \\
 & EvoOcc (ours) & 40.3\dlt{6.9} & 34.5\dlt{3.2} & 41.5\dlt{2.2} \\
\midrule
\multirow{3}{*}{\begin{tabular}{@{}l@{}}OPUS-T /\\ OPUS-V2-L\end{tabular}}
 & Fast System   & 33.2 & 28.7 & 38.4 \\
 & Slow System   & 38.6 & 34.0 & 44.0 \\
 & EvoOcc (ours) & 37.5\dlt{4.3} & 32.2\dlt{3.5} & 40.3\dlt{1.9} \\
\midrule
\multirow{3}{*}{\begin{tabular}{@{}l@{}}ALOcc-2D-mini /\\ OPUS-V2-L\end{tabular}}
 & Fast System   & 33.4 & 31.3 & 39.3 \\
 & Slow System   & 38.6 & 34.0 & 44.0 \\
 & EvoOcc (ours) & 39.6\dlt{6.2} & 33.8\dlt{2.5} & 41.6\dlt{2.3} \\
\bottomrule
\end{tabular}
\end{table}

\begin{table}[t]
\centering
\caption{Results for EvoOcc under different slow system invocation intervals in streaming evolution, on the ALOcc-2D-mini / ALOcc-3D pair. Interval denotes the time between consecutive slow system invocations. The standalone fast and slow systems are listed for reference.}
\label{tab:slow_interval}
\footnotesize
\begin{tabular}{@{}lccccc@{}}
\toprule
  &
  & \multicolumn{3}{c}{Accuracy $\uparrow$}
  & Cost $\downarrow$ \\
\cmidrule(lr){3-5} \cmidrule(l){6-6}
Method & Interval (s) & mIoU & mIoU\textsubscript{D} & RayIoU & TFLOPs/scene \\
\midrule
Fast System & -- & 33.4 & 31.3 & 39.3 & \phantom{0}7.1 \\
Slow System & -- & 38.3 & 35.6 & 43.7 & 15.4 \\
\midrule
\multirow{6}{*}{EvoOcc (ours)}
  & \phantom{0}0.5 & 40.3 & 34.5 & 41.5 & 23.6 \\
  & \phantom{0}2.0 & 39.7 & 34.0 & 40.9 & 13.2 \\
  & \phantom{0}5.0 & 38.9 & 33.5 & 40.4 & 11.0 \\
  & 10.0 & 37.4 & 32.6 & 39.9 & 10.3 \\
  & 15.0 & 36.8 & 32.4 & 39.8 & 10.3 \\
  & 20.0 & 35.6 & 31.7 & 39.4 & \phantom{0}9.9 \\
\bottomrule
\end{tabular}
\end{table}

\begin{table}[t]
\centering
\caption{System-level results for EvoOcc in streaming evolution on the ALOcc-2D-mini / ALOcc-3D pair with a 5.0\,s slow system invocation interval. E2E cost covers the fast system, the slow system, and all EvoOcc operations; latency and FLOPs are averaged per frame, and Mem. denotes peak memory. EvoOcc overhead is the incremental cost of EvoOcc itself.}
\label{tab:system_cost}
\footnotesize
\setlength{\tabcolsep}{4.5pt}
\begin{tabular}{@{}l*{8}{c}@{}}
\toprule
  & \multicolumn{3}{c}{Accuracy $\uparrow$}
  & \multicolumn{3}{c}{E2E cost $\downarrow$}
  & \multicolumn{2}{c}{EvoOcc overhead $\downarrow$} \\
\cmidrule(lr){2-4} \cmidrule(lr){5-7} \cmidrule(l){8-9}
Method
  & mIoU & mIoU\textsubscript{D} & RayIoU
  & Lat.\ (ms) & TFLOPs & Mem.\ (GB)
  & Lat.\ (ms) & TFLOPs \\
\midrule
Fast System   & 33.4 & 31.3 & 39.3 &  63.3 & 0.177 & 0.82 & --   & --    \\
Slow System   & 38.3 & 35.6 & 43.7 & 216.6 & 0.384 & 8.40 & --   & --    \\
EvoOcc (ours) & 38.9 & 33.5 & 40.4 & 101.8 & 0.273 & 8.80 & 23.2 & 0.061 \\
\bottomrule
\end{tabular}
\end{table}

\subsubsection{Experiments in Streaming Evolution Scenarios}

We further evaluate EvoOcc in a more realistic and challenging streaming scenario without additional training. In this scenario, the forward inference latency of the slow system is approximately 0.2\,s, while Occ3D-nuScenes annotates occupancy frames every 0.5\,s. 
Therefore, slow system inference starting from the frame sensed at $t_m$ completes before the next timestamp and its prediction naturally becomes available at $t_n = t_m + 0.5$\,s, incurring a 0.5\,s arrival delay. At $t_n$, EvoOcc recalibrates the historical anchor state using this delayed prediction from $t_m$ and then evolves it to the current timestamp $t_n$ for subsequent state evolution.
Under this strictly causal streaming setting, the results shown in Tab.~\ref{tab:tab2} indicate that EvoOcc consistently improves the fast system across all three representative model pairs by 4.3--6.9 mIoU, 2.5--3.5 mIoU\textsubscript{D}, and 1.9--2.3 RayIoU. Notably, the solid gains in mIoU\textsubscript{D} and RayIoU demonstrate that the dual state evolution can reliably evolve the delayed high-quality slow system features to the current timestamp, effectively compensating for coordinate shifts and scene changes to enhance timely occupancy prediction.

Moreover, since our dual state evolution reliably updates the historical state, it eliminates the necessity to invoke the computationally heavy slow system at every timestamp. To investigate this advantage, we evaluate EvoOcc on a single RTX 3080 Ti GPU using slow system predictions delayed by 0.5\,s while varying the invocation interval from 0.5\,s to 20.0\,s. As shown in Tab.~\ref{tab:slow_interval}, when the slow system is invoked every 0.5\,s, it must compute at every step, resulting in a high computational cost of 23.6 TFLOPs per scene. When the interval is extended to 5.0\,s, the slow system only predicts at 0\,s, 5.0\,s, 10.0\,s, and 15.0\,s, which decreases the total number of slow invocations to merely four and slashes the computation to 11.0 TFLOPs. Despite such substantially sparser inputs, the performance of EvoOcc degrades gracefully, retaining 38.9 mIoU which still surpasses the standalone slow system. 
Even under the extreme 20.0\,s setting, where the slow system is invoked only once at the beginning of the scene, EvoOcc continues to outperform the timely fast system across all three metrics. This demonstrates the stability of our continuous time formulation in suppressing severe error accumulation while retaining useful information over long horizons. Such sparse execution makes it practical to incorporate larger, more accurate slow models into resource-constrained deployment.


The 5.0\,s interval provides a favorable balance between perception performance and computational cost. Under this operating schedule, we evaluate the complete asynchronous pipeline, with end-to-end measurements accounting for the fast system, the slow system, and all EvoOcc operations. As shown in Tab.~\ref{tab:system_cost}, the complete EvoOcc system achieves an average end-to-end latency of 101.8\,ms compared with 216.6\,ms for the slow system, while reducing the average per-frame computation from 0.384 to 0.273\,TFLOPs. Furthermore, EvoOcc itself introduces only 23.2\,ms extra latency and 0.061\,TFLOPs per frame, demonstrating that our framework can achieve a more favorable accuracy-latency balance with limited incremental overhead.

\subsubsection{Experiments under Irregular Fast System Arrival}

\begin{table}[t]
\centering
\caption{Results for EvoOcc under irregular fast system arrival rates. We evaluate EvoOcc without retraining under fixed 3\,Hz, alternating 3/6\,Hz, and random $\{2,3,6\}$\,Hz intervals. Random results are reported as mean$\pm$std over 10 seeds.}
\label{tab:tab5}
\footnotesize
\setlength{\tabcolsep}{5pt}
\begin{tabular}{@{}ll*{3}{Q}@{}}
\toprule
Fast / Slow pair & Fast rate & mIoU & mIoU\textsubscript{D} & RayIoU \\
\midrule
\multirow{3}{*}{\begin{tabular}{@{}l@{}}ALOcc-2D-mini~\cite{chen2025alocc} /\\ ALOcc-3D\end{tabular}}
 & Fixed 3\,Hz      & 38.7 & 33.3 & 40.5 \\
 & Alt.\ 3/6\,Hz    & 38.2 & 33.3 & 40.5 \\
 & Rand.\ 2/3/6\,Hz & 38.6\std{0.21} & 33.3\std{0.04} & 40.5\std{0.02} \\
\midrule
\multirow{3}{*}{\begin{tabular}{@{}l@{}}OPUS-T~\cite{wang2024opus} /\\ OPUS-V2-L\end{tabular}}
 & Fixed 3\,Hz      & 36.5 & 31.4 & 39.2 \\
 & Alt.\ 3/6\,Hz    & 35.9 & 31.2 & 39.1 \\
 & Rand.\ 2/3/6\,Hz & 36.4\std{0.31} & 31.4\std{0.12} & 39.2\std{0.08} \\
\midrule
\multirow{3}{*}{\begin{tabular}{@{}l@{}}ALOcc-2D-mini /\\ OPUS-V2-L\end{tabular}}
 & Fixed 3\,Hz      & 38.8 & 33.5 & 41.2 \\
 & Alt.\ 3/6\,Hz    & 38.4 & 33.3 & 41.1 \\
 & Rand.\ 2/3/6\,Hz & 38.7\std{0.24} & 33.5\std{0.10} & 41.1\std{0.02} \\
\bottomrule
\end{tabular}
\end{table}

To simulate realistic inference conditions where upstream sensor frequencies may fluctuate, we further evaluate EvoOcc under irregular fast system arrivals. Specifically, in the offline evolution scenario, we test different fast prediction rates, including fixed 3\,Hz, alternating 3/6\,Hz, and random $\{2,3,6\}$\,Hz, while keeping the slow system delay fixed at 2.0\,s. As shown in Tab.~\ref{tab:tab5}, the perception performance remains exceptionally stable across all three different fast-slow system pairs, with mIoU\textsubscript{D} varying by at most 0.2 points and RayIoU by at most 0.1 points.

Overall, these results strongly support our continuous time controlled dynamics formulation. By evolving states according to the continuous elapsed time rather than rigid discrete steps, EvoOcc can seamlessly adapt to irregular control signals, demonstrating robust stability and adaptability for unpredictable real-world deployment scenarios.

\subsection{Ablation Study}
\label{sec:ablation}


\begin{table}[t]
\centering
\caption{Ablation study in streaming evolution on the ALOcc-2D-mini / ALOcc-3D pair. w/o control increment replaces $\Delta U_{t_n}$ with the elapsed time $\Delta t_n$; learned fusion replaces scene state evolution by directly fusing fast and warped delayed slow logits. Module costs exclude the fast and slow systems.}
\label{tab:ablation}
\footnotesize
\setlength{\tabcolsep}{5pt}
\begin{tabular}{@{}lccccc@{}}
\toprule
 & \multicolumn{3}{c}{Accuracy $\uparrow$}
 & \multicolumn{2}{c}{Module cost $\downarrow$} \\
\cmidrule(lr){2-4} \cmidrule(l){5-6}
Method & mIoU & mIoU\textsubscript{D} & RayIoU & GFLOPs & Latency (ms) \\
\midrule
Fast System & 33.4 & 31.3 & 39.3 & -- & -- \\
Slow System & 38.3 & 35.6 & 43.7 & -- & -- \\
\midrule
w/o control increment             & 39.9 & 34.2 & 41.1 & 60.2 & 22.9 \\
w/o $\mathcal{L}_{\mathrm{term}}$ & 40.6 & 34.4 & 40.4 & 60.8   & 23.2   \\
Learned fusion (convolution)      & 39.8 & 34.0 & 41.1 & 62.9 & 22.1 \\
Learned fusion (attention)        & 38.4 & 32.9 & 40.2 & 55.3 & 25.2 \\
EvoOcc (ours)                     & 40.3 & 34.5 & 41.5 & 60.8 & 23.2 \\
\bottomrule
\end{tabular}
\end{table}

We conduct an ablation study to assess the contribution of our core components in EvoOcc, as shown in Tab.~\ref{tab:ablation}. In the w/o control increment ablation, we retain the dual-state architecture but replace the control increment $\Delta U_{t_n}$ with the elapsed time $\Delta t_n$, changing the update equation in scene state evolution to $H^{e}_{t_n} = \tilde{H}^{e}_{t_n} + \mathcal{G}^{e}_{\theta} (\tilde{H}^{e}_{t_n}, X^{e}_{t_n}) \Delta t_n$. The control increment $\Delta U_{t_n}$ in Eq.~\eqref{eq:motion_decoupled_euler} measures scene variations and adjusts the evolution magnitude accordingly. Specifically, dynamic objects produce larger variations due to their motion, while static objects produce smaller variations due to stationary properties, enabling $\Delta U_{t_n}$ to adjust these different objects adaptively. When we replace the control increment $\Delta U_{t_n}$ with the elapsed time $\Delta t_n$, this driving signal becomes spatially uniform across all objects, weakening the model's sensitivity to dynamic scene changes.

The learned-fusion baselines instead replace the entire scene state evolution and directly fuse the current fast logits with the warped delayed slow logits using a residual dilated-convolution block and window attention, respectively. To ensure a fair comparison, we appropriately adapt these learnable fusion modules to our framework under comparable computational overheads, as supported by the module FLOPs reported in Tab.~\ref{tab:ablation}. Benefiting from effective updates to static objects, the three baselines achieve performance close to ours, while our EvoOcc still consistently outperforms them in overall accuracy across all semantic and geometric metrics.

\begin{wraptable}[8]{r}{0.42\linewidth}
\centering
\vspace{-1.2em}
\caption{Surface-geometry diagnostics ($\downarrow$) for the ray-termination loss on the ALOcc-2D-mini / ALOcc-3D pair.}
\label{tab:tab7}
\vspace{-0.6em}
\footnotesize
\setlength{\tabcolsep}{5pt}
\begin{tabular}{@{}lcc@{}}
\toprule
Metric & w/o $\mathcal{L}_{\mathrm{term}}$ & EvoOcc \\
\midrule
Extra occ.\ (\%)    & 4.30 & 3.01 \\
Extra length (m)    & 0.72 & 0.65 \\
Extra thickness (m) & 0.15 & 0.10 \\
\bottomrule
\end{tabular}
\vspace{-0.8em}
\end{wraptable}

Although removing the ray-termination loss yields slightly higher mIoU, it noticeably degrades RayIoU. This result indicates that some voxel-wise gains come from more aggressive occupancy completion rather than more accurate surface placement~\cite{liu2024fully}. To further show this effect, we evaluate three surface-geometry metrics on the ALOcc-2D-mini / ALOcc-3D pair, which quantify extra occupancy ratio, length, and thickness outside the ground-truth geometry. As shown in Tab.~\ref{tab:tab7}, EvoOcc achieves the lower values on all three metrics, indicating that $\mathcal{L}_{\rm term}$ effectively suppresses aggressive occupancy completion and improves surface-level geometry. Overall, with all components enabled, EvoOcc improves both voxel-level prediction quality and fine geometric fidelity, yielding a favorable accuracy-latency trade-off.

\section{Conclusion and Limitations}
\label{conclusion}
In this work, we present EvoOcc, a fast-slow evolutionary framework for real-time 3D semantic occupancy prediction. EvoOcc exploits the complementary strengths of deployment-friendly and performance-oriented occupancy predictors to achieve a more favorable accuracy-latency balance, while keeping the upstream architectures unchanged. By modeling  evolutionary occupancy prediction as a continuous time controlled dynamical system and decoupling the evolution process into ego state and scene state evolution, EvoOcc effectively handles asynchronous prediction arrivals. We demonstrate that EvoOcc consistently improves fast system predictions while maintaining much lower latency than the slow system regardless of the combination of fast-slow predictor pairs and the prediction arrival patterns. These findings suggest that delayed high-quality predictions can also benefit real-time predictions when their complementary information is properly evolved, offering a promising alternative pathway toward deploying 3D occupancy systems.

\paragraph{Broader Impact.} EvoOcc can improve real-time 3D occupancy prediction while preserving low response latency, benefiting downstream motion forecasting and planning under practical autonomous driving scenarios. Beyond autonomous driving, the fast-slow evolutionary perspective may also inspire other real-time intelligence systems.

\paragraph{Limitations.} Despite these advantages, several limitations remain. EvoOcc still depends on the prediction quality of the underlying fast and slow systems. In addition, we currently focus on asynchronous arrival caused by inference latency and irregular prediction arrivals, without explicitly modeling uncertainty from signal transmission, or communication delay in real deployment.

\begin{ack}
This research did not receive any specific grant from funding agencies in the public, commercial, or not-for-profit sectors. The authors declare that they have no known competing financial interests or personal relationships that could have appeared to influence the work reported in this paper.
\end{ack}

\bibliographystyle{unsrtnat}
\bibliography{references}

\newpage
\appendix

\section{Derivation of the Ego-Frame Controlled Dynamics}
\label{app:ego_derivation}

This appendix derives the ego-frame transport identities in Eq.~\eqref{eq:ego_transport_identities} and the discrete control increment used in Sec.~\ref{sec:method:evolution}. The derivation is written for a scalar feature channel; for multi-channel evolving states and fast features, the same argument applies channel-wise. Spatial arguments are shown only when needed.

\paragraph{Coordinate transform and transport term.}
Let $\Omega^w$ and $\Omega$ denote the world and ego-frame spatial domains. Let $T_t:\Omega\rightarrow\Omega^w$ be the ego-to-world transform at time $t$. For an ego-frame grid location $p\in\Omega$, denote its corresponding world location by $P_t=T_t(p)$. The ego-frame fields are the pullbacks of the world-frame fields:
\begin{equation}
H_t^e(p)=H_t^w(P_t),
\qquad
X_t^e(p)=X_t^w(P_t).
\label{eq:app_pullback}
\end{equation}
Differentiating the state pullback at a fixed ego-grid location $p$ gives
\begin{equation}
\frac{d}{dt}H_t^e(p)
=
\partial_t H_t^w(P_t)
+
\nabla_w H_t^w(P_t)\cdot \dot P_t .
\label{eq:app_hidden_chain}
\end{equation}
Let $J_t(p)=\partial T_t(p)/\partial p$ be the spatial Jacobian. Since $\nabla_e H_t^e(p)=J_t(p)^\top\nabla_w H_t^w(P_t)$, we define the ego-grid transport velocity under the sign convention of the main text as
\begin{equation}
u_t(p):=-J_t(p)^{-1}\dot P_t .
\label{eq:app_transport_velocity}
\end{equation}
Then Eq.~\eqref{eq:app_hidden_chain} can be rewritten as
\begin{equation}
\partial_t H_t^w(P_t)
=
\partial_t H_t^e(p)+u_t(p)\cdot\nabla_e H_t^e(p).
\label{eq:app_hidden_transport_derivative}
\end{equation}
Applying the same chain-rule argument to $X_t$ yields
\begin{equation}
dH_t^w
=
dH_t^e+(u_t\cdot\nabla H_t^e)\,dt,
\qquad
dX_t^w
=
dX_t^e+(u_t\cdot\nabla X_t^e)\,dt,
\label{eq:app_ego_transport_identities}
\end{equation}
where $\nabla$ denotes the ego-grid gradient. This gives Eq.~\eqref{eq:ego_transport_identities}: the world-frame variation at fixed physical locations is expressed as the sum of the temporal variation observed on the ego grid and the apparent transport induced by ego motion.

\paragraph{Ego-frame controlled dynamics.}
The world-reference dynamics in Eq.~\eqref{eq:world_cde} are defined as
\begin{equation}
dH_t^w=\mathcal G_\theta^w(H_t^w,X_t^w)dX_t^w .
\label{eq:app_world_cde}
\end{equation}
Substituting Eq.~\eqref{eq:app_ego_transport_identities} into Eq.~\eqref{eq:app_world_cde}, and denoting the ego-coordinate counterpart of the local spatial operator by $\mathcal G_\theta^e$, gives
\begin{equation}
dH_t^e+(u_t\cdot\nabla H_t^e)\,dt
=
\mathcal G_\theta^e(H_t^e,X_t^e)
\Bigl(dX_t^e+(u_t\cdot\nabla X_t^e)\,dt\Bigr).
\label{eq:app_ego_cde_expanded}
\end{equation}
We collect the transport-aligned fast feature variation into a control increment,
\begin{equation}
dU_t:=dX_t^e+(u_t\cdot\nabla X_t^e)\,dt,
\qquad
\dot U_t=\partial_tX_t^e+u_t\cdot\nabla X_t^e .
\label{eq:app_control_definition}
\end{equation}
Thus, the ego-frame dynamics become
\begin{equation}
dH_t^e+(u_t\cdot\nabla H_t^e)\,dt
=
\mathcal G_\theta^e(H_t^e,X_t^e)dU_t .
\label{eq:app_ego_controlled_dynamics}
\end{equation}
The left-hand side accounts for deterministic coordinate transport of the state, while $dU_t$ measures the fast-feature variation after aligning the previous feature field to the same physical locations.

\paragraph{Discrete approximation of the control increment.}
Consider an online interval $[t_{n-1},t_n]$. The continuous control accumulated over this interval is
\begin{equation}
\Delta U_{t_n}
:=
U_{t_n}-U_{t_{n-1}}
=
\int_{t_{n-1}}^{t_n}
\left(\partial_tX_t^e+u_t\cdot\nabla X_t^e\right)dt .
\label{eq:app_delta_u_integral}
\end{equation}
Let $\chi_{t_{n-1}\leftarrow t_n}(p)$ be the ego-grid location at $t_{n-1}$ that corresponds to the same physical point as the current ego-grid location $p$ at $t_n$. The backward warping operator induced by ego motion evaluates the previous fast feature at this transported location:
\begin{equation}
\tilde X_{t_n}^e(p)
=
W_{\mathrm{ego}}(X_{t_{n-1}}^e)(p)
\approx
X_{t_{n-1}}^e\bigl(\chi_{t_{n-1}\leftarrow t_n}(p)\bigr).
\label{eq:app_warped_x}
\end{equation}
By the first-order semi-Lagrangian approximation along the transport trajectory,
\begin{equation}
\int_{t_{n-1}}^{t_n}
\left(\partial_tX_t^e+u_t\cdot\nabla X_t^e\right)dt
\approx
X_{t_n}^e(p)-X_{t_{n-1}}^e\bigl(\chi_{t_{n-1}\leftarrow t_n}(p)\bigr).
\label{eq:app_semilagrangian}
\end{equation}
Combining Eq.~\eqref{eq:app_delta_u_integral}--\eqref{eq:app_semilagrangian} yields the discrete increment used in the main text:
\begin{equation}
\Delta U_{t_n}
\approx
X_{t_n}^e-W_{\mathrm{ego}}(X_{t_{n-1}}^e)
=
X_{t_n}^e-\tilde X_{t_n}^e .
\label{eq:app_delta_u_discrete}
\end{equation}
Therefore, $\Delta U_{t_n}$ is not a raw ego-grid difference between two consecutive fast features; it is a transport-aligned approximation of the continuous control increment over $[t_{n-1},t_n]$.

\section{Additional Implementation Details}

Unless otherwise specified, all latency results reported in the main paper are measured on a single NVIDIA RTX 3080 Ti GPU. This runtime measurement is independent of the training hardware, where EvoOcc is trained on two NVIDIA RTX 5090 GPUs. All upstream fast and slow systems use officially released checkpoints. 


\paragraph{Voxelization and ego-motion warping.}
We follow the standard Occ3D-nuScenes~\cite{tian2023occ3d} voxelization setting. The point-cloud range is set to
$[-40.0, -40.0, -1.0, 40.0, 40.0, 5.4]$, the voxel size is
$[0.4, 0.4, 0.4]$, and the resulting occupancy grid size is
$200 \times 200 \times 16$. For the ego-motion warping operator $W_{\rm ego}$, we use the official ego poses provided by nuScenes to construct the ego-to-global transformation at each timestamp. Given two timestamps $t_{n-1}$ and $t_n$, we compute their relative ego motion from the corresponding ego-to-global matrices and warp the state and fast feature to the current ego grid. This uses only the standard pose and calibration information provided by nuScenes~\cite{caesar2020nuscenes} and introduces no additional external localization or calibration source. For the warped delayed slow baseline, out-of-boundary regions after warping are completed with the corresponding fast system predictions at the current timestamp.

\paragraph{Efficient realization of the controlled term.}
Eq.~\eqref{eq:motion_decoupled_euler} writes the scene state evolution in the form of a controlled update, where $\mathcal G_\theta^e(\tilde H^e_{t_n}, X^e_{t_n})$ maps the control increment to a state increment. A fully general realization would require predicting a dense matrix-valued operator for each voxel, which is computationally expensive for 3D occupancy grids. We therefore use a factorized realization of the controlled term.

Specifically, we augment the transport-aligned control increment with the elapsed time,
\begin{equation}
    \Delta \bar U_{t_n}
    =
    \operatorname{Concat}(\Delta t_n, \Delta U_{t_n}),
    \qquad
    \Delta t_n = t_n - t_{n-1},
    \label{eq:app_augmented_control}
\end{equation}
where $\Delta t_n$ is broadcast over the spatial grid. A bias-free point-wise linear projection $\psi_\theta$ maps $\Delta \bar U_{t_n}$ to the hidden-state dimension, while a convolutional local operator $\phi_\theta$ predicts state-dependent modulation coefficients from the transported state and the current fast feature:
\begin{equation}
    \psi_\theta(\Delta \bar U_{t_n}) = W\Delta \bar U_{t_n}.
    \label{eq:app_factorized_components}
\end{equation}
The controlled increment is then computed by applying the predicted modulation coefficients as a diagonal matrix to the projected augmented control:
\begin{equation}
    \delta H^e_{t_n}
    =
    \operatorname{Diag}\!\left(
    \phi_\theta(\tilde H^e_{t_n}, X^e_{t_n})
    \right)
    \psi_\theta(\Delta \bar U_{t_n})
    =
    \operatorname{Diag}\!\left(
    \phi_\theta(\tilde H^e_{t_n}, X^e_{t_n})
    \right)
    W\Delta \bar U_{t_n}.
    \label{eq:app_factorized_increment}
\end{equation}
The Euler update becomes
\begin{equation}
    H^e_{t_n}
    =
    \tilde H^e_{t_n}
    +
    \delta H^e_{t_n}.
    \label{eq:app_factorized_euler}
\end{equation}
This factorization can be interpreted as a restricted matrix-valued interaction between the current state-feature context and the augmented control increment. Since $\Delta \bar U_{t_n}$ contains both the fast-feature change $\Delta U_{t_n}$ and the elapsed time $\Delta t_n$, the hidden state can still receive a time-dependent update even when the short-term fast system change is small. Thus, the practical implementation provides a lightweight time-dependent component while keeping the main formulation driven by the fast system prediction stream. Therefore, the implementation remains consistent with the controlled dynamics in Eq.~\eqref{eq:ego_transport_cde}, while avoiding the cost of explicitly constructing dense voxel-wise matrices.

\paragraph{Ray sampling for the ray-termination loss.}
For $\mathcal L_{\rm term}$, we use the same virtual LiDAR ray directions as the RayIoU~\cite{liu2024fully}. Specifically, each LiDAR origin emits $R=14040$ rays generated from predefined pitch angles and $1^\circ$ azimuth intervals. For each supervised timestamp, we keep at most $K=8$ LiDAR origins from the past and future ego poses provided by the sweep annotations. Along each ray, we uniformly sample 50 points with a step size of $0.4\,{\rm m}$, corresponding to depths $d_i=(i+0.5)\times0.4\,{\rm m}$ and covering approximately $20\,{\rm m}$. The free-space probability at each sampled point is obtained from the predicted occupancy logits by softmax followed by trilinear grid sampling. The termination window in Eq.~\eqref{eq:ray_termination_loss} is defined around the ground-truth surface hit depth $d_r^\ast$ as
\[
    \mathcal W_r=\{i\,:\, |d_i-d_r^\ast|\leq 0.4\,{\rm m}\}.
\]
This implementation makes $\mathcal L_{\rm term}$ directly aligned with the ray-based surface evaluation while keeping the supervision restricted to local neighborhoods around visible surface hits.

\section{Additional Experiments}

\begin{table}[t]
\centering
\caption{Per-class results on the Occ3D-nuScenes~\cite{tian2023occ3d} dataset in the offline evolution scenario. The slow system is evaluated directly at the current timestamp and reported only as a reference.}
\label{tab:tab9}
\scriptsize
\setlength{\tabcolsep}{1.8pt}
\begin{tabular}{@{}lc@{\hspace{5pt}}ccccccccccccccccc@{}}
\toprule
Method & mIoU & \rotatebox{90}{others} & \rotatebox{90}{barrier} & \rotatebox{90}{bicycle} & \rotatebox{90}{bus} & \rotatebox{90}{car} & \rotatebox{90}{cons. veh.} & \rotatebox{90}{motorcycle} & \rotatebox{90}{pedestrian} & \rotatebox{90}{tfc. cone} & \rotatebox{90}{trailer} & \rotatebox{90}{truck} & \rotatebox{90}{drive. surf.} & \rotatebox{90}{other flat} & \rotatebox{90}{sidewalk} & \rotatebox{90}{terrain} & \rotatebox{90}{manmade} & \rotatebox{90}{vegetation} \\
\midrule
\multicolumn{19}{@{}l}{\textit{ALOcc-2D-mini / ALOcc-3D}} \\
\hspace{0.6em}Fast System & 33.38 & 13.18 & 46.24 & 26.06 & 40.45 & 46.89 & 22.58 & 26.61 & 26.64 & 29.73 & 25.93 & 35.53 & 59.82 & 41.29 & 42.10 & 36.08 & 24.70 & 23.68 \\
\hspace{0.6em}Slow System & 38.29 & 14.27 & 49.09 & 31.48 & 42.11 & 50.26 & 27.74 & 31.63 & 29.82 & 36.30 & 32.98 & 38.73 & 64.80 & 44.37 & 46.93 & 42.45 & 35.57 & 32.32 \\
\hspace{0.6em}EvoOcc (ours) & 39.31 & 14.27 & 49.43 & 26.70 & 43.36 & 50.94 & 23.54 & 27.80 & 28.59 & 32.64 & 30.37 & 38.29 & 76.26 & 45.52 & 53.06 & 52.84 & 39.30 & 35.36 \\
\midrule
\multicolumn{19}{@{}l}{\textit{OPUS-T / OPUS-V2-L}} \\
\hspace{0.6em}Fast System & 33.19 & 10.70 & 39.83 & 21.26 & 39.75 & 45.25 & 23.40 & 21.83 & 17.81 & 19.24 & 27.49 & 33.20 & 71.61 & 37.11 & 45.14 & 43.59 & 33.80 & 33.18 \\
\hspace{0.6em}Slow System & 38.60 & 12.97 & 49.53 & 27.98 & 43.95 & 51.49 & 24.36 & 29.72 & 26.63 & 34.58 & 30.20 & 37.96 & 72.10 & 40.84 & 48.78 & 46.93 & 41.04 & 37.18 \\
\hspace{0.6em}EvoOcc (ours) & 36.88 & 11.36 & 45.72 & 24.83 & 41.67 & 49.33 & 24.81 & 25.28 & 20.38 & 22.04 & 29.99 & 36.18 & 74.42 & 42.89 & 51.82 & 50.95 & 38.41 & 36.78 \\
\midrule
\multicolumn{19}{@{}l}{\textit{ALOcc-2D-mini / OPUS-V2-L}} \\
\hspace{0.6em}Fast System & 33.38 & 13.18 & 46.24 & 26.06 & 40.45 & 46.89 & 22.58 & 26.61 & 26.64 & 29.73 & 25.93 & 35.53 & 59.82 & 41.29 & 42.10 & 36.08 & 24.70 & 23.68 \\
\hspace{0.6em}Slow System & 38.60 & 12.97 & 49.53 & 27.98 & 43.95 & 51.49 & 24.36 & 29.72 & 26.63 & 34.58 & 30.20 & 37.96 & 72.10 & 40.84 & 48.78 & 46.93 & 41.04 & 37.18 \\
\hspace{0.6em}EvoOcc (ours) & 39.12 & 14.05 & 48.86 & 26.89 & 42.89 & 50.50 & 23.87 & 27.94 & 28.28 & 32.00 & 31.00 & 37.58 & 75.77 & 44.88 & 52.03 & 51.86 & 40.32 & 36.24 \\
\bottomrule
\end{tabular}
\end{table}

\begin{table}[t]
\begin{minipage}[b]{0.43\linewidth}
\caption{Offline evolution results on the SurroundOcc-nuScenes~\cite{wei2023surroundocc} with the ALOcc-2D-mini / ALOcc-3D pair.}
\label{tab:surroundocc}
\end{minipage}\hfill
\begin{minipage}[b]{0.53\linewidth}
\caption{Results for EvoOcc across five training seeds in offline evolution on the Occ3D-nuScenes dataset, reported as mean$\pm$std over seeds 0--4.}
\label{tab:training_seeds}
\end{minipage}

\begin{minipage}[t]{0.43\linewidth}
\vspace{0pt}\centering\footnotesize
\setlength{\tabcolsep}{4pt}
\renewcommand{\arraystretch}{1.1}
\begin{tabular}{@{}lccc@{}}
\toprule
Method & mIoU & mIoU\textsubscript{D} & IoU \\
\midrule
Fast System         & 21.5 & 19.5 & 31.5 \\
Slow System         & 24.0 & 21.7 & 34.7 \\
Warped delayed slow & 18.9 & 14.3 & 32.3 \\
EvoOcc (ours)       & 23.5 & 20.9 & 35.9 \\
\bottomrule
\end{tabular}
\end{minipage}\hfill
\begin{minipage}[t]{0.53\linewidth}
\vspace{0pt}\centering\footnotesize
\setlength{\tabcolsep}{4pt}
\begin{tabular}{@{}l*{3}{Q}@{}}
\toprule
Fast / Slow pair & mIoU & mIoU\textsubscript{D} & RayIoU \\
\midrule
\begin{tabular}{@{}l@{}}ALOcc-2D-mini /\\ ALOcc-3D\end{tabular}
 & 39.28\std{0.04} & 33.69\std{0.09} & 40.55\std{0.04} \\
\midrule
\begin{tabular}{@{}l@{}}OPUS-T /\\ OPUS-V2-L\end{tabular}
 & 36.84\std{0.06} & 31.51\std{0.04} & 39.45\std{0.02} \\
\bottomrule
\end{tabular}
\end{minipage}
\end{table}

\paragraph{Per-class IoU analysis.}
As shown in Tab.~\ref{tab:tab9}, we further provide a per-class IoU analysis on Occ3D-nuScenes. Across all three fast-slow system pairs, EvoOcc consistently achieves higher IoU than the fast system baselines for every semantic category, with improvements covering both large static regions, such as driveable surface, sidewalk, and challenging dynamic or small-object categories, such as bicycle, pedestrian and traffic cone. In several categories, EvoOcc also approaches or even surpasses the current-time slow system reference. These results demonstrate that our dual state evolution is effective, as it uses the fast system prediction stream to evolve the historical anchor state, thereby handling coordinate shifts and genuine scene changes.

\paragraph{Transfer beyond Occ3D-nuScenes.}
We also evaluate EvoOcc on SurroundOcc-nuScenes~\cite{wei2023surroundocc}. As shown in Tab.~\ref{tab:surroundocc}, EvoOcc improves the fast system by 2.0 mIoU and 1.4 mIoU\textsubscript{D}, demonstrating better overall semantic occupancy estimation and dynamic scene modeling. Its gains of 4.6 mIoU and 6.6 mIoU\textsubscript{D} over warped delayed slow further show that the improvement cannot be explained by ego-motion warping alone. These results provide additional evidence that our dual state evolution framework remains effective under the different occupancy annotation protocol.

\paragraph{Statistical results across multiple training runs.}
We independently train EvoOcc using five random seeds (0–4) on two fast-slow system pairs. The random seed controls the parameter initialization of EvoOcc and stochastic operations during training. The frozen upstream systems, data split, and all other training settings remain unchanged across runs. 
As shown in Tab.~\ref{tab:training_seeds}, the standard deviation is at most 0.09 across all metrics. The mean results also closely match the single-seed results reported in Tab.~\ref{tab:tab1}. These results demonstrate that EvoOcc remains stable across independent training runs and that its improvements do not depend on a particular random seed.

\paragraph{Surface-geometry diagnostics.}
We introduce three surface-geometry diagnostics to quantify aggressive occupancy completion around the ground-truth surface. The extra occupancy ratio measures redundant occupied predictions in front of the ground-truth surface, reflecting premature occupancy before the true surface hit. The extra length measures the occupied extent behind the ground-truth surface, indicating whether the prediction continues to fill the space after the true surface. The extra thickness measures redundant vertical thickness for flat classes, including driveable surface, other flat, sidewalk and terrain, capturing whether thin ground-like regions are predicted as thick or elevated structures. Higher values indicate more severe over-completion and less accurate surface geometry. As shown in Tab.~\ref{tab:tab7}, EvoOcc achieves the lower values on all three diagnostics, showing that $\mathcal{L}_{\rm term}$ effectively suppresses aggressive occupancy completion and improves surface-level geometry.


\newpage
\section*{NeurIPS Paper Checklist}

\begin{enumerate}

\item {\bf Claims}
    \item[] Question: Do the main claims made in the abstract and introduction accurately reflect the paper's contributions and scope?
    \item[] Answer: \answerYes{}
    \item[] Justification: We have stated the challenge, our method, and our contributions in both the abstract and introduction.

\item {\bf Limitations}
    \item[] Question: Does the paper discuss the limitations of the work performed by the authors?
    \item[] Answer: \answerYes{}
    \item[] Justification: We have discussed the limitations in Sec.~\ref{conclusion}.

\item {\bf Theory assumptions and proofs}
    \item[] Question: For each theoretical result, does the paper provide the full set of assumptions and a complete (and correct) proof?
    \item[] Answer: \answerNA{}
    \item[] Justification: We do not present formal theoretical results. We provide mathematical derivations for the ego-frame controlled dynamics and the discrete control increment in Appendix~\ref{app:ego_derivation}.

\item {\bf Experimental result reproducibility}
    \item[] Question: Does the paper fully disclose all the information needed to reproduce the main experimental results of the paper to the extent that it affects the main claims and/or conclusions of the paper (regardless of whether the code and data are provided or not)?
    \item[] Answer: \answerYes{}
    \item[] Justification: We have discussed the training implementation details and the dataset construction details in Sec.~\ref{experiments:4.1}. We will make our source code publicly available.

\item {\bf Open access to data and code}
    \item[] Question: Does the paper provide open access to the data and code, with sufficient instructions to faithfully reproduce the main experimental results, as described in supplemental material?
    \item[] Answer: \answerNo{}
    \item[] Justification: We have provided implementation details, and we will make our source code publicly available.

\item {\bf Experimental setting/details}
    \item[] Question: Does the paper specify all the training and test details (e.g., data splits, hyperparameters, how they were chosen, type of optimizer, etc.) necessary to understand the results?
    \item[] Answer: \answerYes{}
    \item[] Justification: We have specified the training implementation details and the dataset construction details in Sec.~\ref{experiments:4.1}.

\item {\bf Experiment statistical significance}
    \item[] Question: Does the paper report error bars suitably and correctly defined or other appropriate information about the statistical significance of the experiments?
    \item[] Answer: \answerNo{}
    \item[] Justification: Following the common deterministic evaluation protocol in 3D occupancy prediction, we fix the random seed to 0 and disable other sources of randomness when possible. Therefore, the main benchmark results are reported under a controlled deterministic setting without error bars or statistical significance tests. For the random fast-arrival setting, we additionally report mean and standard deviation over 10 random seeds in Tab.~\ref{tab:tab5}.

\item {\bf Experiments compute resources}
    \item[] Question: For each experiment, does the paper provide sufficient information on the computer resources (type of compute workers, memory, time of execution) needed to reproduce the experiments?
    \item[] Answer: \answerYes{}
    \item[] Justification: We have specified the experimental compute resource details in Sec.~\ref{experiments:4.1}.
    
\item {\bf Code of ethics}
    \item[] Question: Does the research conducted in the paper conform, in every respect, with the NeurIPS Code of Ethics \url{https://neurips.cc/public/EthicsGuidelines}?
    \item[] Answer: \answerYes{}
    \item[] Justification: We have read and adhered to the NeurIPS Code of Ethics in conducting our research.

\item {\bf Broader impacts}
    \item[] Question: Does the paper discuss both potential positive societal impacts and negative societal impacts of the work performed?
    \item[] Answer: \answerYes{}
    \item[] Justification: In Sec.~\ref{conclusion}, we have discussed the broader impacts of our work.
    
\item {\bf Safeguards}
    \item[] Question: Does the paper describe safeguards that have been put in place for responsible release of data or models that have a high risk for misuse (e.g., pre-trained language models, image generators, or scraped datasets)?
    \item[] Answer: \answerNA{}
    \item[] Justification: There are no such risks associated with our work.

\item {\bf Licenses for existing assets}
    \item[] Question: Are the creators or original owners of assets (e.g., code, data, models), used in the paper, properly credited and are the license and terms of use explicitly mentioned and properly respected?
    \item[] Answer: \answerYes{}
    \item[] Justification: We properly cite the original papers for all datasets and baseline models in Sec.~\ref{sec:method} and Sec.~\ref{experiments}. Specifically, the nuScenes dataset is used under the CC BY-NC-SA 4.0 license, the Occ3D benchmark is used under its MIT and CC BY-NC-SA 4.0 terms, and the open-source codebases and baseline models (e.g., built upon PyTorch under the BSD license and MMDetection3D under the Apache-2.0 license) are used in strict compliance with their respective terms of use solely for non-commercial academic research.

\item {\bf New assets}
    \item[] Question: Are new assets introduced in the paper well documented and is the documentation provided alongside the assets?
    \item[] Answer: \answerNA{}
    \item[] Justification: This paper does not introduce new datasets or physical assets, and we will release the source code alongside full documentation.

\item {\bf Crowdsourcing and research with human subjects}
    \item[] Question: For crowdsourcing experiments and research with human subjects, does the paper include the full text of instructions given to participants and screenshots, if applicable, as well as details about compensation (if any)?
    \item[] Answer: \answerNA{}
    \item[] Justification: Our paper does not involve crowdsourcing experiments or research with human subjects.

\item {\bf Institutional review board (IRB) approvals or equivalent for research with human subjects}
    \item[] Question: Does the paper describe potential risks incurred by study participants, whether such risks were disclosed to the subjects, and whether Institutional Review Board (IRB) approvals (or an equivalent approval/review based on the requirements of your country or institution) were obtained?
    \item[] Answer: \answerNA{}
    \item[] Justification: Our work does not involve human subjects.

\item {\bf Declaration of LLM usage}
    \item[] Question: Does the paper describe the usage of LLMs if it is an important, original, or non-standard component of the core methods in this research? Note that if the LLM is used only for writing, editing, or formatting purposes and does \emph{not} impact the core methodology, scientific rigor, or originality of the research, declaration is not required.
    \item[] Answer: \answerNA{}
    \item[] Justification: LLMs are not used as an important, original, or non-standard component of the core methods in this research. They are used only for writing, editing, formatting, and coding assistance, which does not affect the core methodology, scientific rigor, or originality of the research.

\end{enumerate}

\end{document}